\documentclass[sigconf]{acmart}
\usepackage{graphicx}
\usepackage{algorithm}
\usepackage{algpseudocode}
\usepackage{amsmath}

\usepackage{amssymb}
\usepackage{booktabs}
\usepackage{hyperref}
\usepackage{xcolor}
\usepackage{soul}
\usepackage{multirow}
\definecolor{revisionbg}{RGB}{232, 242, 254}
\sethlcolor{revisionbg}
\renewcommand{\hl}[1]{#1}
\AtBeginDocument{%
  }

\copyrightyear{2026}
\acmYear{2026}
\setcopyright{cc}
\setcctype{by}
\acmConference[BCB '26]{17th ACM International Conference on Bioinformatics, Computational Biology and Health Informatics}{June 30-July 03, 2026}{Rende (CS), Italy}
\acmBooktitle{17th ACM International Conference on Bioinformatics, Computational Biology and Health Informatics (BCB '26), June 30-July 03, 2026, Rende (CS), Italy}
\acmDOI{10.1145/3807503.3819480}
\acmISBN{979-8-4007-2653-8/2026/06}

\newcommand{\pms}[1]{{\scriptsize$\pm$\,#1}}

\begin{document}

\title{Efficient Imputation for Patch-Based Missing Single-Cell Data via Cluster-Regularized Optimal Transport}

\author{Yuyu Liu}
\affiliation{%
  \institution{Department of Computer Science, Stony Brook University}
  \city{Stony Brook}
  \state{NY}
  \country{USA}
}

\author{Jiannan Yang}
\affiliation{%
  \institution{Department of Computer Science, Stony Brook University}
  \city{Stony Brook}
  \state{NY}
  \country{USA}
}

\author{Ziyang Yu}
\affiliation{%
  \institution{Department of Computer Science, Emory University}
  \city{Atlanta}
  \state{GA}
  \country{USA}
}

\author{Weishen Pan}
\affiliation{%
  \institution{Department of Population Health Sciences, Cornell University}
  \city{Ithaca}
  \state{NY}
  \country{USA}
}

\author{Fei Wang}
\affiliation{%
  \institution{Department of Computer Science and Engineering, Cornell University}
  \city{Ithaca}
  \state{NY}
  \country{USA}
}

\author{Tengfei Ma}
\authornote{Corresponding author.}
\affiliation{%
  \institution{Department of Biomedical Informatics, Stony Brook University}
  \city{Stony Brook}
  \state{NY}
  \country{USA}
}

\renewcommand{\shortauthors}{Liu et al.}

\begin{abstract}
Missing data in single-cell sequencing datasets poses significant challenges for extracting meaningful biological insights. However, existing imputation approaches, which often assume uniformity and data completeness, struggle to address cases with large patches of missing data. In this paper, we present \textbf{CROT} (\textbf{C}luster-\textbf{R}egularized \textbf{O}ptimal \textbf{T}ransport), an optimal transport-based imputation algorithm designed to handle patch-based missing data in tabular formats. Our approach effectively captures the underlying data structure in the presence of significant missingness. Notably, it achieves superior imputation accuracy while significantly reducing runtime, demonstrating its scalability and efficiency for large-scale datasets. This work introduces a robust solution for imputation in heterogeneous, high-dimensional datasets with structured data absence, addressing critical challenges in both biological and clinical data analysis. Our code is available on GitHub\footnote{\scriptsize \url{https://github.com/yuyuliu11037/CROT}.}.

\end{abstract}


\maketitle

\section{Introduction}

\begin{figure}[ht!]
    \centering
    \includegraphics[width=0.48\textwidth]{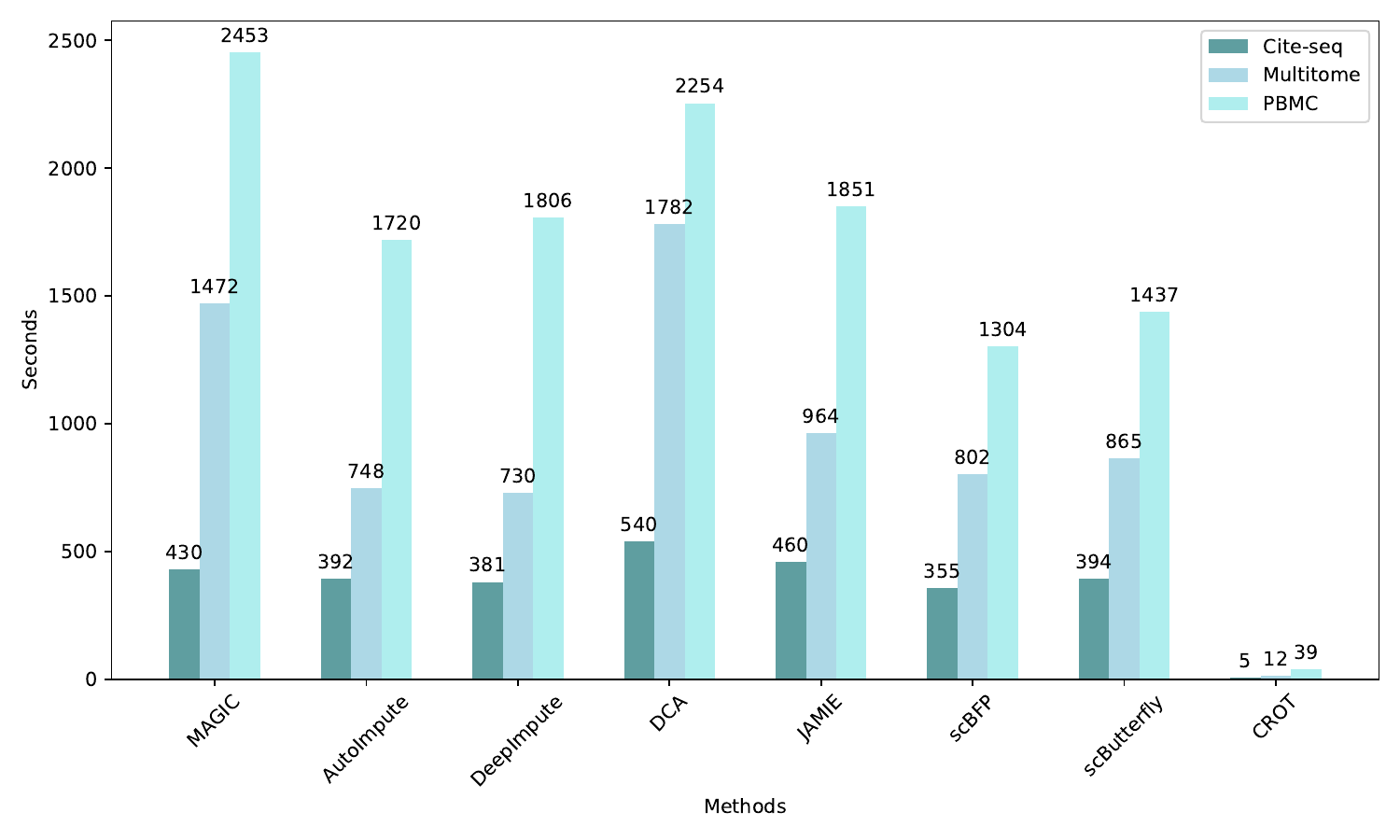}
    \caption{Runtime comparison on three datasets. Calculated as the total time of model initialization, model training, and inference on the dataset (if the method is model-dependent), or the total time of iterative interpolation on the target dataset (if the method is not model-dependent). The final results are truncated to integers.}
    \label{fig:runnint_time}
\end{figure}

Single-cell sequencing technologies have revolutionized our ability to profile gene expression at the resolution of individual cells, uncovering diverse cell types and states within complex tissues. However, these datasets are notoriously sparse, containing many zero values due to both biological reasons (true lack of gene expression) and technical limitations (insufficient mRNA capture or sequencing depth). Such “dropout” events can severely hinder downstream analysis and diminish the power of single-cell studies if left unaddressed~\cite{jia_accounting_2017}. In practice, this means important biological insights—like identifying rare cell subpopulations or accurately mapping developmental trajectories—may be obscured by noise. Imputation of missing or zeroed-out values is therefore critical to recover meaningful biological signals and enable robust analysis of cellular heterogeneity.

A growing number of methods have been developed to tackle dropout imputation in single-cell RNA-seq data. Early approaches such as MAGIC~\cite{dijk_recovering_2018} diffuse information across similar cells to smooth out zeros, while scImpute~\cite{li_accurate_2018} fits mixture models to distinguish true zero expression from technical dropouts. Bayesian methods like SAVER~\cite{huang_saver_2018} borrow information across genes to infer likely expression values. More recently, deep learning techniques have emerged: multi-modal frameworks like JAMIE~\cite{kalafut_jamie_nodate} use variational autoencoders to impute one modality from another. However, most existing methods make strong assumptions about data completeness or uniform missingness and tend to borrow information only from local similarities (neighboring cells or genes). This can lead to over-smoothing of the data (blurring true biological distinctions) and loss of natural cell-to-cell variability. Crucially, many methods struggle when faced with patch-based missing data—the extreme scenario where entire blocks of features (such as whole gene sets or entire assay modalities) are absent for a subset of the dataset. Such situations can arise, for example, if a technical failure causes all cells from one experimental batch to lack a particular measurement (e.g., all ADT protein tags missing in one CITE-seq batch, or all gene expression values missing for a condition in a multi-omic experiment). Traditional dropout-focused algorithms are not equipped to handle these large, structured absences, since they assume each feature is at least partially observed across the dataset. 

To address these challenges, we propose \textbf{C}luster-\textbf{R}egularized \textbf{O}ptimal \textbf{T}ransport (CROT). Our approach is motivated by two key insights: (1) When one subset of data is missing entire features, a powerful way to infer those missing values is to align the distribution of the incomplete data with that of a related complete dataset. (2) Simply aligning global distributions might ignore the internal structure that is crucial in biology. Specifically, the presence of distinct cell subpopulations or clusters (e.g. cell types, states). Therefore, CROT augments the OT mapping with a cluster regularization term that preserves the similarity of cluster centroids (representative cell-type expression profiles) between the complete and incomplete data. By explicitly enforcing that the imputed data recapitulates the same cell-type structure as the reference data, we maintain inter-cell-type specificity and avoid collapsing distinct cell identities. The biological relevance of this strategy is clear. For example, if T cells and B cells form separate clusters in a fully observed dataset, our method strives to ensure that after imputation, T cells and B cells in the previously incomplete dataset are still well-separated and each cluster’s prototype expression remains accurate. Our contributions can be summarized as follows:

\begin{itemize}
    \item We propose a cluster-regularized optimal transport imputation method that effectively captures the intricate relations within single-cell data batches, demonstrating high accuracy in imputing dropout events.
    \item Our method achieves not only superior imputation performance but also demonstrates fast convergence, making it highly efficient for large-scale single-cell datasets. Figure~\ref{fig:runnint_time} previews this efficiency gain: CROT completes imputation in seconds where baselines require many minutes.
    \item We evaluate CROT on three real-world single-cell datasets (CITE-seq, Multiome, PBMC) under simulated patch-based modality dropout. Our results indicate that CROT outperforms state-of-the-art imputation methods in both accuracy and computational efficiency.
\end{itemize}

\section{Related Work}
In this section, we review related work, including existing OT-based and non-OT-based imputation methods for single-cell and general data.

\subsection{Single-cell Sequencing Data Imputation}
Single-cell sequencing data frequently encounters dropout events, leading to numerous zeros that obstruct downstream analyses. To mitigate this issue, a variety of computational and statistical methods have been proposed. Notable approaches include MAGIC~\cite{dijk_recovering_2018}, which utilizes data diffusion to smooth out dropouts; scImpute~\cite{li_accurate_2018}, which distinguishes true zeros from dropouts through a mixture model; and SAVER~\cite{huang_saver_2018}, which applies a Bayesian approach to impute missing values while considering gene expression variability. Additionally, scGNN~\cite{wang_scgnn_2021} employs graph neural networks to account for cellular neighborhood relationships, while ALRA~\cite{linderman_zero-preserving_2022} utilizes adaptive low-rank autoregressive models to capture temporal dependencies in gene expression profiles. Although these methods have demonstrated efficacy in enhancing clustering, visualization, and gene expression analysis, they often struggle under conditions of extensive missing data and may not fully leverage the inherent structural relationships within the data. Recent multimodal probabilistic models have further improved cross-modality inference in single-cell data. totalVI~\cite{gayoso_joint_2021} jointly models RNA expression and protein abundance (ADT) using a variational inference framework that explicitly accounts for technical noise and protein background. MultiVI~\cite{ashuach_multivi_2023} extends this framework to integrate RNA expression with chromatin accessibility (ATAC) by learning a shared latent representation across modalities. These models enable modality translation and data integration even when one modality is missing, making them widely used baselines for multimodal imputation tasks.

\subsection{Optimal Transport in Data Imputation}
Optimal transport (OT) has been widely employed for data imputation across various domains. For instance,~\cite{muzellec_missing_2020} proposed an OT-based loss function for imputing missing values, demonstrating its efficacy across multiple datasets. Similarly,~\cite{wu_jointly_2023} introduced GIT, a generative imputation model that leverages OT for multi-view data imputation, achieving notable performance improvements in multi-view scenarios. In the realm of single-cell multi-modal data,~\cite{alatkar_cmot_2023} developed CMOT (Cross-Modality Optimal Transport), which utilizes OT to align cells across different modalities and infer missing data. Additionally,~\cite{klein_generative_2023} presented the GENOT framework, which employs entropic OT to predict cellular responses and translate across data modalities, thereby addressing complex data integration challenges. Optimal transport has also been widely studied for domain adaptation, where class-aware regularization is incorporated to preserve semantic structure during distribution alignment~\cite{courty_optimal_2017}. These methods enforce that samples from the same class are transported together. Another class of OT-based multimodal alignment methods includes SCOT~\cite{demetci_scot_2022} and its improved variant SCOTv2~\cite{demetci_scotv2_2022}, which leverage Gromov-Wasserstein optimal transport to align single-cell datasets across modalities while preserving intra-domain geometry.  

Despite their successes, these methods often overlook the critical link between data integrity and downstream tasks, such as cell-type clustering, that depend on accurate and biologically meaningful imputation. \hl{Our approach is conceptually related but differs in a fundamental way. In supervised domain adaptation, class labels are known and fixed. The regularization simply prevents labeled groups from mixing during transport. In our setting, the cluster structure is latent and must be discovered jointly with the imputation. Because entire feature blocks are absent, the initial clustering of incomplete data is unreliable, and cluster quality depends on imputation quality, which in turn depends on the cluster regularization. This mutual dependence requires an alternating optimization that co-evolves clustering and imputation, a dynamic absent in supervised class-aware OT. Moreover, our goal is data imputation rather than domain transfer, which changes how the transport plan is used: we directly optimize the missing entries via gradient descent on the Sinkhorn objective, rather than mapping samples across domains.}


\begin{figure*}[t]
    \centering
    \includegraphics[width=1.0\textwidth]{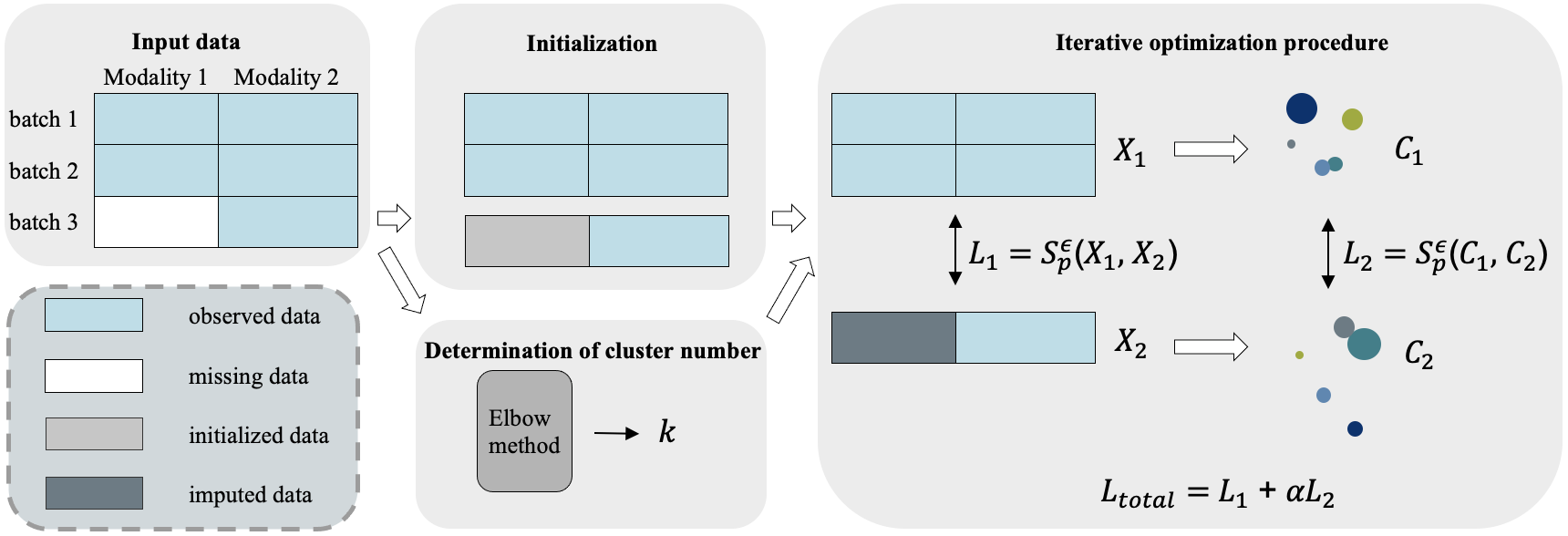}
    \caption{An illustration of our framework. Assume that all modality 1 data from batch 3 is missing (where ``modality" may vary by dataset). The process begins by initializing the missing data and using observed data from batches 1 and 2 to determine an optimal number of clusters. The following iterative optimization procedure is then applied. In each iteration, clustering with $k$ clusters is performed to assign a class label to each row in $X_1$ and $X_2$, respectively. Next, centroids $C_1$ and $C_2$ are calculated by averaging the rows belonging to each class. Subsequently, the Sinkhorn divergence between ($X_1, X_2$) and ($C_1, C_2$) is computed to form the total loss. This procedure continues until the total loss converges.}
    \label{fig:method}
\end{figure*}

\subsection{Patchwork Learning}
In recent work, Patchwork Learning (PL) has emerged as a compelling approach to handling heterogeneous data sources, particularly in healthcare contexts where data is distributed across multiple sites with varying modalities.~\citet{rajendran_patchwork_2024} introduce PL as a framework that enables the integration of disparate datasets, even when certain sites lack complete modalities, by leveraging bridging modalities or shared features across sites. This approach aligns closely with our missing data patch setting, where partial features from one or more sites are unavailable. By addressing the challenge of incomplete data across distributed networks, PL facilitates more robust and generalizable machine learning models, capable of imputing missing data and ensuring data privacy. 

\section{Preliminary}
Optimal transport (OT) theory provides a powerful tool for comparing probability distributions by solving the problem of moving mass between distributions at minimal cost. Given two probability distributions $\mu$ and $\nu$ supported on $X$, the
$p$-Wasserstein distance is defined as

\begin{equation}
W_p(\mu,\nu)
=
\left(
\inf_{\gamma \in \Pi(\mu,\nu)}
\int_{X \times X}
\|x-y\|^p \, d\gamma(x,y)
\right)^{1/p}
\end{equation}

where $\Pi(\mu,\nu)$ denotes the set of couplings with marginals
$\mu$ and $\nu$. This distance, though geometrically meaningful, is computationally expensive, particularly in high dimensions.

To alleviate the computational burden, entropic regularization adds
an entropy penalty to the OT objective:

\begin{equation}
W^{\epsilon}_p(\mu,\nu)
=
\left(
\inf_{\gamma \in \Pi(\mu,\nu)}
\int_{X \times X}
\|x-y\|^p d\gamma(x,y)
+
\epsilon\, \mathrm{KL}(\gamma \,\|\, \mu \otimes \nu)
\right)^{1/p}
\end{equation}

where \(\text{KL}(\cdot \| \cdot)\) is the Kullback-Leibler divergence between \(\gamma\) and the independent coupling \(\mu \otimes \nu\), and \(\epsilon > 0\) controls the strength of the regularization. The added entropy facilitates faster convergence via iterative algorithms, notably the Sinkhorn-Knopp algorithm.

The resulting quantity, known as the Sinkhorn divergence, approximates the original Wasserstein distance while offering significant computational speed-ups:
\begin{equation}
    S^\epsilon_p(\mu, \nu) := W_p^\epsilon(\mu, \nu) - \frac{1}{2} \left( W_p^\epsilon(\mu, \mu) + W_p^\epsilon(\nu, \nu) \right)
\end{equation}

This divergence retains the essential properties of the Wasserstein distance but benefits from the computational efficiency introduced by the regularization, making it particularly well-suited for large-scale applications where OT-based techniques are needed to compare distributions.

\section{Methodology}

In this section, we present the details of CROT for imputing missing values in tabular data and analyze its computational complexity. Our framework is shown in Figure \ref{fig:method}.

\begin{algorithm}
\caption{Cluster-regularized Optimal Transport}
\label{alg:crot}
\raggedright
\textbf{Input:} $X_1 \in \mathbb{R}^{m_1 \times n}$, $X_2 \in \mathbb{R}^{m_2 \times n}$, $\epsilon,T,\alpha,l > 0$,  missing columns $S$ in $X_2$, number of clusters  $k$\\
\textbf{Output:} Completed data $\hat{X}_2$\\
\textbf{Initialization:} 
\begin{algorithmic}[1]
    \For{$j \in S$}
        \For{$i = 1, \ldots, m_2$}
            \State $\hat{X}_2[i,j] \gets \mathrm{mean}({X}_1[:,j]) + \epsilon_{ij}$, 
            
            \ \ \ where $\epsilon_{ij} \sim \mathcal{N}(0, 1)$
        \EndFor
    \EndFor
\end{algorithmic}
\textbf{Optimization:} 
\begin{algorithmic}[1]
\For{$t = 1, \ldots, T$}
    \State Sample two sets $K$, $L$ of $l$ indices from $X_1$ and $\hat{X}_2$, respectively
    \State $(\mathcal{C}_{1K},\hat{\mathcal{C}}_{2L})\gets \text{Clustering}_k(X_{1K}, \hat{X}_{2L})$
    \State $\mathcal{L} \gets S^\epsilon_p(X_{1K}, \hat{X}_{2L}) + \alpha S^\epsilon_p (\mathcal{C}_{1K},\hat{\mathcal{C}}_{2L})$
    \State $\hat{X}^{(imp)}_{2L} \gets \hat{X}^{(imp)}_{2L} - \text{Adam}(\nabla_{\hat{X}^{(imp)}_{2L}}\mathcal{L})$
\EndFor
\end{algorithmic}
\end{algorithm}

\subsection{Cluster-regularized Optimal Transport}
\label{sec:method}


To impute missing data, we first consider information transfer at the data level. Let $X_1$ and $X_2$ be two sets of tabular data, where $X_1 \in \mathbb{R}^{m_1 \times n}$ is the set of $m_1$ observed data, and $X_2 \in \mathbb{R}^{m_2 \times n}$ is the set containing missing values to be imputed. For empirical distributions with uniform weights, the coupling
$\gamma$ becomes a transport matrix
$\gamma \in \mathbb{R}^{m_1 \times m_2}$. We can simplify the Wasserstein distance between these two sets of discrete samples $X_1$ and $X_2$ to 

\begin{equation}
W_p(X_1, X_2)
=
\left(
\inf_{\gamma}
\sum_{i,j}
\|X_1[i,:] - X_2[j,:]\|^p
\, \gamma_{ij}
\right)^{1/p}
\end{equation}

where we use the Euclidean $L_2$ distance as the cost function $c$, and the marginals of transference plan $\gamma$ are uniform distributions.

Consequently, the discrete version of Sinkhorn divergence between $X_1$ and $X_2$, denoted by $S_p^\epsilon (X_1, X_2)$, constitutes the main part of the total loss function. Minimizing this distance maps features in $X_1$ to features in $X_2$, thereby imputing missing values in $X_2$.

However, this method ignores the consistency of clustering structures between different sets of data. For $X_1$ and $X_2$, we expect not only the alignment between the data sample distributions but also the alignment between cluster distributions. For example, for different sets of single cell sequencing data in the same domain, we may want their cell-type prototypes also to remain similar. 
Therefore, we incorporate cluster regularization by computing the Wasserstein distance between the cluster centers of the complete data and the missing data. The cluster centers are obtained from the average of entries within each class, calculated along the rows of $X_1$ and $X_2$.

In order to obtain effective clusters, we use the Elbow method to heuristically determine the optimal number $k$ of clusters in $X_1$. Based on the similarity assumption of $X_1$ and $X_2$, we regard it as an approximation of the number of clusters in $X_2$. At each iteration, we perform clustering with $k$ clusters on $X_1$ and $X_2$, where the cluster centers are given by:
\begin{align*}
    c_s(X_1) &:= \frac{1}{|C_{1,s}|} \sum_{i \in C_s} X_{1}[i,:]
    \\
     c_t(X_2) &:= \frac{1}{|C_{2,t}|} \sum_{j \in C_{2,t}} X_{2}[j,:]
\end{align*}
where $C_{1,s}$ is the set of indices assigned to cluster $s$ in $X_1$, and similar for $C_{2,t}$. These cluster centers are then used to compute the Wasserstein distance between the complete data (from $X_1$) and the imputed data (from $X_2$). Let the two sets of cluster centers computed from $X_1, X_2$ be $\mathcal{C}_1$ and $\mathcal{C}_2$, respectively. Finally, we define the cluster-regularized OT loss as follows:
\begin{equation}
    L_{\mathrm{CROT}}(X_1,X_2) = S^\epsilon_p(X_1, X_2) + \alpha S^\epsilon_p (\mathcal{C}_1,\mathcal{C}_2)
\end{equation}

The second term is a regularization term that encourages the cluster centers of the imputed data to align closely with the complete data. The constant $\alpha>0$ controls the trade-off between these two objectives. In all experiments, we fix the OT hyperparameters to $\epsilon = 0.1$ and
$N_{\mathrm{SK}} = 100$ Sinkhorn iterations per OT evaluation, and optimize the
imputed entries using Adam with learning rate $\eta = 10^{-2}$. Unless early stopping
is triggered, the outer optimization is run for $T = 150$ iterations using mini-batches
of l = 3000 rows from $X_1$ and $\hat{X}_2$. Since the overall distribution of the two data matrices is relatively similar, the relationship between each cluster is also similar, resulting in no order of magnitude difference between calculating the Sinkhorn distance between the entire data and using only the center points. Therefore, no additional adjustments are needed to $\alpha$. Finally, we illustrate the overall optimization strategy of CROT in Algorithm \ref{alg:crot}.

\paragraph{Differentiability of the clustering step.}
The clustering operation used in CROT (k-means) involves a discrete assignment step that is inherently non-differentiable. Specifically, the mapping from data points to cluster indices is defined through an $\arg\min$ operation over distances to centroids, which prevents gradients from propagating through the assignment variables. In our optimization procedure, we therefore treat the cluster assignments as fixed within each iteration. After performing k-means clustering on the sampled subsets of $X_1$ and $\hat{X}_2$, the assignments are held constant while computing cluster centroids and evaluating the cluster-regularized OT loss.

Under this scheme, gradients are propagated only through the centroid computation and the Sinkhorn divergence terms. Since cluster centroids are defined as averages of assigned samples, they remain differentiable with respect to the underlying data entries in $\hat{X}_2$. The assignment step itself is recomputed at the beginning of each iteration, resulting in an alternating optimization procedure that interleaves discrete clustering updates with continuous gradient-based updates of the imputed matrix.

This alternating strategy is commonly used in cluster-regularized learning objectives. While it does not provide gradients through the discrete assignment variables, in practice it yields stable optimization because the centroid updates provide smooth signals for the transport loss while the clustering step periodically refreshes the structural constraints.


\subsection{Complexity Analysis}
Let $T$ denote the number of outer optimization iterations and
$T_s$ denote the number of Sinkhorn iterations used to compute
the transport plan. The complexity analysis of the algorithm is divided into the following two main phases:

\paragraph{Phase 1: Initialization}

During the initialization process, the algorithm iterates over each missing column indexed by \( j \) in \( S \) and each row \( i \) in \( X_2 \), where \( X_2[i,j] \) is updated based on the mean of corresponding entries in \( X_1 \). Computing this mean requires \( O(m_1) \) time, where \( m_1 \) is the number of rows in \( X_1 \). Given that there are \( |S| \) missing columns in \( X_2 \), the time complexity for initialization is \( O(|S| \cdot m_1) \). Meanwhile, The Elbow method is used to determine the optimal number of clusters $k$ for a k-means clustering algorithm using $X_1$. It involves running k-means $K$ with different values of $k$ and calculating the within-cluster sum of squares (WCSS) for each $k$. The time complexity of the k-means algorithm costs \(O(l\cdot k\cdot i)\), where $i$ is a constant that represents the number of iterations; and running for $K$ times results in \(O(K\cdot i\cdot m_1\cdot k\cdot n)\), which could be simplified to \(O(m_1\cdot n)\) since $K$ is typically a small number compared to $m_1$. Therefore, the total time complexity of phase 1 is \(O(m_1\cdot n)\)

\paragraph{Phase 2: Iterative Optimization}

The main loop runs for \( T \) iterations. In each iteration, two sets of \( l \) indices are sampled from matrices \( X_1 \) and \( X_2 \), costing \( O(l) \) time. For each sampled set with shape \(l\times n\), k-means algorithm is applied to assign a label for each row, costing \(O(l\cdot k\cdot i)\). The Sinkhorn algorithm requires $T_s$ iterations, resulting in
a time complexity of $O(T_s \cdot l^2)$, where \( T \) is the number of iterations, dominated by operations on the \( l \times l \) transport matrix. The Adam optimizer adds a minor overhead of \( O(l \cdot n) \) for the parameter update step. However, since \( l^2 \) dominates, the overall time complexity per iteration remains \( O(T \cdot l^2) \).

\begin{table*}[htbp]
\centering
\caption{Results of missing data recovery on CITE-seq, Multiome, and PBMC. Each value is reported as mean ± standard deviation over five random batch pairs.}
\label{tab:similarity_sc}
\begin{tabular}{lccccccccc}
\toprule
\textbf{Dataset} & \multicolumn{3}{c}{\textbf{CITE-seq}} & \multicolumn{3}{c}{\textbf{Multiome}} & \multicolumn{3}{c}{\textbf{PBMC}} \\
\cmidrule(lr){2-4} \cmidrule(lr){5-7} \cmidrule(lr){8-10}
\textbf{Metrics} & \textbf{PCC} & \textbf{MAE} & \textbf{RMSE} & \textbf{PCC} & \textbf{MAE} & \textbf{RMSE} & \textbf{PCC} & \textbf{MAE} & \textbf{RMSE} \\
\midrule
RAW\footnotemark & nan & 3.99\pms{0.03} & 4.10\pms{0.06} & nan & 2.07\pms{0.10} & 2.18\pms{0.07} & nan & 1.30\pms{0.05} & 1.44\pms{0.09} \\
MAGIC & 0.72\pms{0.11} & 1.03\pms{0.06} & 1.19\pms{0.04} & 0.17\pms{0.01} & 1.08\pms{0.05} & 1.19\pms{0.07} & 0.62\pms{0.04} & 0.92\pms{0.09} & 1.04\pms{0.15} \\
AutoImpute & 0.70\pms{0.12} & 0.56\pms{0.03} & \textbf{0.62}\pms{0.03} & 0.09\pms{0.00} & 4.70\pms{0.21} & 6.85\pms{0.33} & 0.52\pms{0.02} & 0.93\pms{0.04} & 0.99\pms{0.05} \\
DeepImpute & 0.54\pms{0.12} & 7.61\pms{0.13} & 12.05\pms{0.54} & 0.12\pms{0.02} & 2.36\pms{0.13} & 3.70\pms{0.19} & 0.50\pms{0.02} & 1.53\pms{0.07} & 1.97\pms{0.09} \\
DCA & 0.60\pms{0.01} & 1.08\pms{0.02} & 1.29\pms{0.06} & 0.09\pms{0.01} & 2.44\pms{0.10} & 2.94\pms{0.12} & 0.53\pms{0.12} & 1.89\pms{0.08} & 2.28\pms{0.10} \\
JAMIE & 0.69\pms{0.02} & 1.09\pms{0.09} & 1.27\pms{0.06} & 0.21\pms{0.03} & 1.80\pms{0.09} & 1.97\pms{0.09} & 0.58\pms{0.09} & 1.23\pms{0.06} & 1.40\pms{0.07} \\
scBFP & 0.58\pms{0.02} & 0.70\pms{0.13} & 0.99\pms{0.04} & 0.06\pms{0.01} & 1.45\pms{0.06} & 1.62\pms{0.06} & 0.56\pms{0.04} & 2.03\pms{0.09} & 2.30\pms{0.10} \\
scButterfly & 0.71\pms{0.02} & 0.94\pms{0.06} & 1.25\pms{0.05} & 0.15\pms{0.02} & 1.69\pms{0.07} & 1.88\pms{0.09} & 0.59\pms{0.05} & 0.93\pms{0.05} & 1.17\pms{0.07} \\
totalVI & 0.74\pms{0.09} & 0.82\pms{0.03} & 0.94\pms{0.07} & 0.14\pms{0.01} & 1.63\pms{0.05} & 2.27\pms{0.11} & 0.57\pms{0.06} & 1.41\pms{0.10} & 1.72\pms{0.06} \\
MultiVI & 0.70\pms{0.02} & 0.99\pms{0.05} & 1.43\pms{0.12} & 0.14\pms{0.03} & 1.32\pms{0.20} & 1.94\pms{0.08} & 0.54\pms{0.07} & 0.88\pms{0.06} & 1.85\pms{0.11} \\
SCOTv2 & 0.73\pms{0.04} & 0.68\pms{0.04} & 1.84\pms{0.17} & 0.18\pms{0.01} & 1.04\pms{0.22} & 1.74\pms{0.23} & 0.55\pms{0.04} & 1.03\pms{0.04} & 1.58\pms{0.03} \\
\bottomrule
CROT & \textbf{0.76}\pms{0.04} & \textbf{0.47}\pms{0.02} & 0.99\pms{0.03} & \textbf{0.22}\pms{0.01} & \textbf{0.59}\pms{0.03} & \textbf{1.18}\pms{0.03} & \textbf{0.64}\pms{0.01} & \textbf{0.75}\pms{0.03} & \textbf{0.88}\pms{0.03} \\
\bottomrule
\end{tabular}
\end{table*}

Combining both phases, the overall time complexity of the algorithm is \( O(m_1\cdot n + T \cdot l^2) \). Since $l$ is a small constant with respect to the number of sampled lines, the overall time complexity increases linearly with the size of $X_1$, making the imputation process fast and efficient. The space complexity is dominated by storing matrices \( X_1 \) and \( X_2 \), resulting in \( O(m_1 \cdot n + T \cdot T_s \cdot l^2) \).

\section{Experiment}
In this section, we evaluate the effectiveness of the proposed method. We first measure the similarity between the original and the imputed data, followed by the cell-type task on several widely used datasets. In addition, we demonstrate the lightweightness of the proposed approach by conducting several imputing efficiency comparison studies.

\subsection{Dataset and Experiment Settings}
\paragraph{Datasets.}
To validate the effectiveness of our method on real-world data, we conduct experiments on three widely used single-cell sequencing datasets. The BMMC-citeseq dataset~\cite{luecken_sandbox_2021} contains RNA and protein expression data from bone marrow mononuclear cells obtained using CITE-seq technology, allowing multi-omics analysis. The BMMC-multiome dataset includes paired measurements of chromatin accessibility and gene expression from BMMC, providing insight into gene regulation at a single-cell level. The PBMC dataset \footnote{\url{https://www.10xgenomics.com/datasets}} offers gene expression profiling of peripheral blood mononuclear cells, which is commonly used for investigating immune cell diversity and function.

\paragraph{Benchmark Construction.}
\hl{Each dataset contains multiple experimental batches: CITE-seq comprises 4 site-level batches (16{,}311, 25{,}171, 32{,}029, and 16{,}750 cells; 90{,}261 total), Multiome comprises 4 site-level batches (17{,}243, 15{,}226, 14{,}556, and 22{,}224 cells; 69{,}249 total), and PBMC comprises 8 donor-level batches, each subsampled to 10{,}000 cells. After preprocessing (Appendix\mbox{~\ref{sec:preprocessing}}), the final feature dimensions are: CITE-seq: 2{,}000 HVGs $+$ ${\sim}$134 ADT proteins; Multiome: 4{,}000 ATAC peaks $+$ 2{,}832 HVGs; PBMC: ${\sim}$1{,}714 HVGs $+$ 228 ADT proteins.

To simulate patch-based modality dropout, we designate one or more batches as the complete reference ($X_1$) and one batch as the incomplete target ($X_2$). In $X_2$, all entries of one modality (e.g., all ADT columns or all GEX columns) are replaced by learnable parameters initialized from the column means of $X_1$ plus Gaussian noise $\epsilon_{ij}\sim\mathcal{N}(0,1)$. The original feature columns are retained in shape, a binary mask marks which entries are missing, and only those masked positions are overwritten by trainable variables that are optimized via Adam through the OT loss
(Algorithm\mbox{~\ref{alg:crot}}). The ground-truth values are held out separately for evaluation. To account for variability, we repeat this procedure across five randomly selected (source, target) batch pairs per dataset, ensuring source and target are disjoint, and report the mean and standard deviation of all metrics.}

\paragraph{Experimental Settings.}
All experiments were conducted on a workstation with an NVIDIA RTX 4090 GPU (24 GB) using Python 3.10 and PyTorch 2.1. Unless otherwise specified, our implementation uses GPU acceleration for matrix operations and optimization steps. For fairness, all baseline methods were executed using their official implementations with the recommended hyperparameters provided by the original authors.

In our experiments, we empirically set the regularization parameter $\alpha = 1$, as discussed in Section \ref{sec:method}.

\subsection{Evaluation Metrics}

\paragraph{Numerical Recovery}
We assess the recovery performance of CROT using Root Mean Square Error (RMSE), Mean Absolute Error (MAE), and Pearson Correlation Coefficient (PCC). RMSE and MAE quantify the numerical discrepancies between the imputed data and the true expression values, where lower RMSE and MAE values signify more accurate imputation. PCC, on the other hand, is a similarity measurement that evaluates the correlation between imputed and true expression data. A PCC value closer to 1 indicates a higher degree of similarity, reflecting the effectiveness of the imputation method employed in this study

\paragraph{Clustering Representation}
To evaluate structural preservation, we employ clustering metrics such as Adjusted Rand Index (ARI), Normalized Mutual Information (NMI), and Purity. These metrics assess the extent to which the imputation retains meaningful clusters by comparing the imputed data to the ground-truth clusters. For qualitative analysis, we use Uniform Manifold Approximation and Projection (UMAP) for dimensionality reduction, allowing visual comparison of the original and imputed data in a lower-dimensional space.

\paragraph{Efficiency}
Additionally, we measure running time to evaluate computational efficiency, providing insight into the scalability of the method.

\begin{table*}[htbp]
\centering
\caption{Results of cell type clustering on CITE-seq, Multiome, and PBMC. Each value is reported as mean ± standard deviation over five random batch pairs.}
\label{tab:cluster_sc}
\begin{tabular}{lccccccccc}
\toprule
\textbf{Dataset} & \multicolumn{3}{c}{\textbf{CITE-seq}} & \multicolumn{3}{c}{\textbf{Multiome}} & \multicolumn{3}{c}{\textbf{PBMC}} \\
\cmidrule(lr){2-4} \cmidrule(lr){5-7} \cmidrule(lr){8-10}
\textbf{Metrics} & \textbf{ARI} & \textbf{NMI} & \textbf{PURITY} & \textbf{ARI} & \textbf{NMI} & \textbf{PURITY} & \textbf{ARI} & \textbf{NMI} & \textbf{PURITY} \\
\midrule
RAW & 0.65\pms{0.02} & 0.67\pms{0.03} & 0.77\pms{0.02} & 0.49\pms{0.02} & 0.62\pms{0.03} & 0.67\pms{0.03} & 0.53\pms{0.06} & 0.73\pms{0.04} & 0.95\pms{0.03} \\
AutoImpute & 0.80\pms{0.01} & 0.81\pms{0.12} & 0.89\pms{0.02} & 0.13\pms{0.02} & 0.33\pms{0.03} & 0.34\pms{0.07} & 0.34\pms{0.02} & 0.44\pms{0.03} & 0.90\pms{0.02} \\
DeepImpute & 0.53\pms{0.02} & 0.67\pms{0.03} & 0.82\pms{0.03} & 0.45\pms{0.02} & 0.58\pms{0.03} & 0.60\pms{0.03} & 0.38\pms{0.02} & 0.51\pms{0.03} & 0.84\pms{0.02} \\
DCA & 0.68\pms{0.02} & 0.72\pms{0.03} & 0.78\pms{0.03} & 0.16\pms{0.02} & 0.19\pms{0.03} & 0.62\pms{0.03} & 0.38\pms{0.02} & 0.49\pms{0.03} & 0.88\pms{0.02} \\
JAMIE & 0.72\pms{0.03} & 0.71\pms{0.07} & 0.78\pms{0.04} & 0.44\pms{0.11} & 0.61\pms{0.08} & 0.62\pms{0.07} & 0.40\pms{0.06} & 0.69\pms{0.03} & 0.89\pms{0.07} \\
scBFP & 0.22\pms{0.07} & 0.57\pms{0.03} & 0.84\pms{0.02} & 0.10\pms{0.01} & 0.37\pms{0.03} & 0.50\pms{0.03} & 0.13\pms{0.01} & 0.72\pms{0.03} & 0.92\pms{0.05} \\
scButterfly & 0.69\pms{0.02} & 0.72\pms{0.03} & 0.81\pms{0.03} & 0.36\pms{0.02} & 0.53\pms{0.03} & 0.63\pms{0.03} & 0.21\pms{0.02} & 0.50\pms{0.03} & 0.88\pms{0.02} \\
totalVI & 0.71\pms{0.01} & 0.69\pms{0.01} & 0.84\pms{0.01} & 0.57\pms{0.02} & 0.66\pms{0.00} & 0.59\pms{0.01} & 0.19\pms{0.06} & 0.34\pms{0.03} & 0.81\pms{0.01} \\
MultiVI & 0.74\pms{0.02} & 0.70\pms{0.00} & 0.88\pms{0.01} & 0.60\pms{0.00} & 0.61\pms{0.02} & 0.64\pms{0.02} & 0.53\pms{0.01} & 0.67\pms{0.01} & 0.88\pms{0.01} \\
SCOTv2 & 0.77\pms{0.01} & 0.71\pms{0.01} & 0.86\pms{0.01} & 0.63\pms{0.01} & 0.59\pms{0.03} & 0.65\pms{0.01} & 0.57\pms{0.00} & 0.63\pms{0.02} & 0.89\pms{0.00} \\
\bottomrule
CROT & \textbf{0.83}\pms{0.01} & \textbf{0.75}\pms{0.02} & \textbf{0.93}\pms{0.01} & \textbf{0.68}\pms{0.01} & \textbf{0.79}\pms{0.02} & \textbf{0.70}\pms{0.02} & \textbf{0.64}\pms{0.01} & \textbf{0.77}\pms{0.02} & \textbf{0.96}\pms{0.01} \\
\bottomrule
\end{tabular}
\end{table*}




\begin{figure*}[htbp]
\centering
\includegraphics[width=\linewidth]{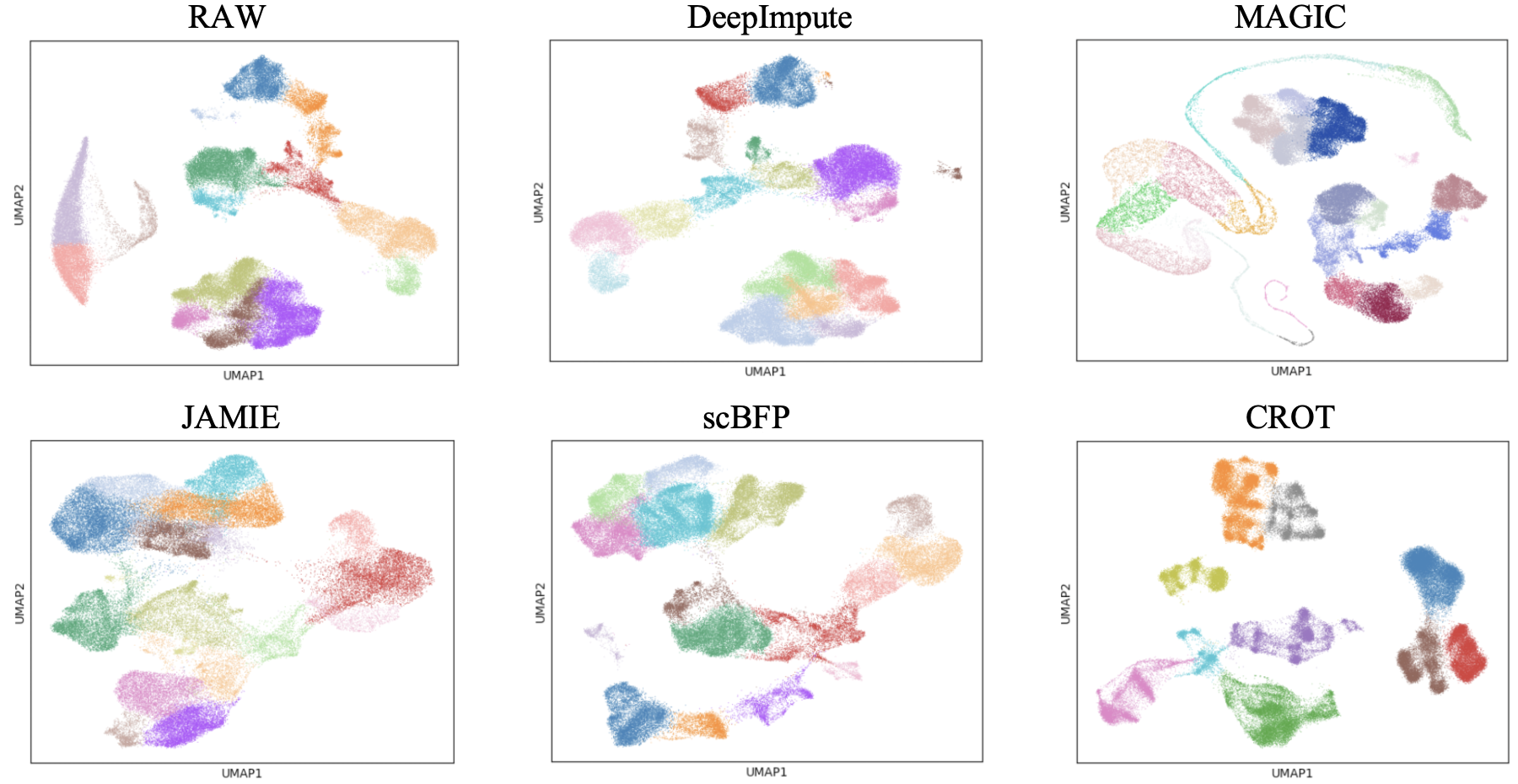} 
\caption{UMAP visualizations comparing clustering quality across different imputation methods. Better distinct clusters of UMAP plots helps with further analysis of single-cell data, such as the presence of rare cell types or intermediate states between well-defined populations.}
\label{fig:umap}
\end{figure*}

\begin{figure*}[htbp]
    \centering
    \includegraphics[width=1.0\textwidth]{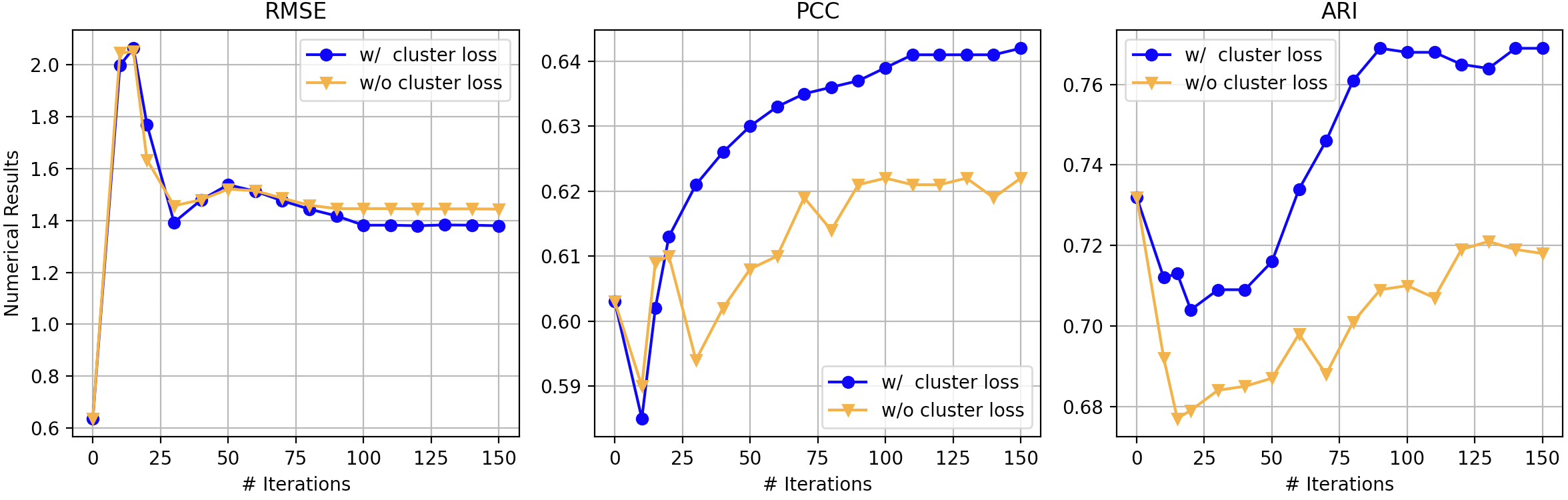}
    \caption{Numerical value of RMSE, PCC and ARI on one missing setting of CITE-seq during iterative imputation process.}
    \label{fig:metrics_evaluation}
\end{figure*}

\subsection{Baselines}
We evaluate the performance of our method against several state-of-the-art models for single-cell sequencing, focusing on tasks such as data recovery and clustering. The selected baselines include both single-modality imputation methods and multimodal integration approaches. Single-modality baselines include MAGIC~\cite{dijk_recovering_2018}, AutoImpute~\cite{talwar_autoimpute_2018}, DeepImpute~\cite{arisdakessian_deepimpute_2019}, scBFP~\cite{lee_single-cell_2024}, and DCA~\cite{eraslan_single-cell_2019}. For multimodal inference, we include JAMIE~\cite{kalafut_jamie_nodate}, scButterfly~\cite{cao_scbutterfly_2024}, totalVI~\cite{gayoso_joint_2021}, and MultiVI~\cite{ashuach_multivi_2023}, which are variational autoencoder–based frameworks for cross-modality prediction. In addition, we include SCOTv2~\cite{demetci_scotv2_2022}, an optimal transport–based multimodal alignment method. Detailed descriptions of these methods are available in Section A.2 of Supplementary Materials. To ensure fair runtime comparison, we followed the original implementations of each baseline method. Methods implemented in deep learning frameworks (e.g., DeepImpute, JAMIE, scButterfly) were executed with GPU acceleration when supported, while CPU-oriented methods (e.g., MAGIC, AutoImpute) were executed on the CPU as required by their official implementations. No algorithmic modifications were introduced.

In addition to these baselines, we also assessed clustering performance on raw data without imputation (denoted as RAW) to demonstrate the necessity of the data imputation. 

\footnotetext{PCC is undefined for RAW: missing entries are constant-filled (column means), so the imputed vector has zero variance and Pearson correlation is not defined.}

\subsection{Missing Data Recovering}
To test the authenticity of the data estimated by our method, we compared the correlation and similarity between the imputed data and the real data. The experimental results on single-cell data are displayed in Table \ref{tab:similarity_sc}. Our method achieves the highest PCC and lowest MAE values across all datasets, indicating superior imputation accuracy compared to other methods. 

\subsection{Clustering Analysis}

\subsubsection{Cell Type and Phenotype Clustering}
Identifying cell types and phenotypes from diverse populations of single cells and patients, respectively, relies on effective clustering techniques, which group entities based on similarities, such as gene expression or clinical features. However, the presence of missing data can interfere with clustering performance. Thus we first apply all imputation methods to get the imputed data and then evaluate the clustering effectiveness using Leiden algorithm. The results in Table \ref{tab:cluster_sc} show that raw data (without imputation) clustering is consistently inferior to imputed data clustering, implying the importance and necessity of imputation. Among the imputation methods, CROT achieves the best performance in cell type clustering on CITE-seq, Multiome, and PBMC datasets, with the highest Adjusted Rand Index (ARI) and Purity, such as 0.83 and 0.93 on CITE-seq, and 0.64 and 0.96 on PBMC. 

\subsubsection{Visualization of Imputed Data}

 In addition, we also provide the UMAP plots of imputed data by several methods on Multiome data in Figure \ref{fig:umap}. Compared with the graph for raw data, where the clusters appear less distinct and more dispersed, all methods show gradual improvements in cluster separation. Notably, our method demonstrates significantly improved clustering with more distinct clusters than others, indicating that our imputation method better preserves biological heterogeneity and enhances the data's underlying structure. Therefore, our approach more effectively captures the relationships between cells, resulting in more biologically meaningful clusters.

\subsubsection{Marker Protein Expression Analysis}
\begin{figure*}[t]
  \centering
  \includegraphics[width=\textwidth]{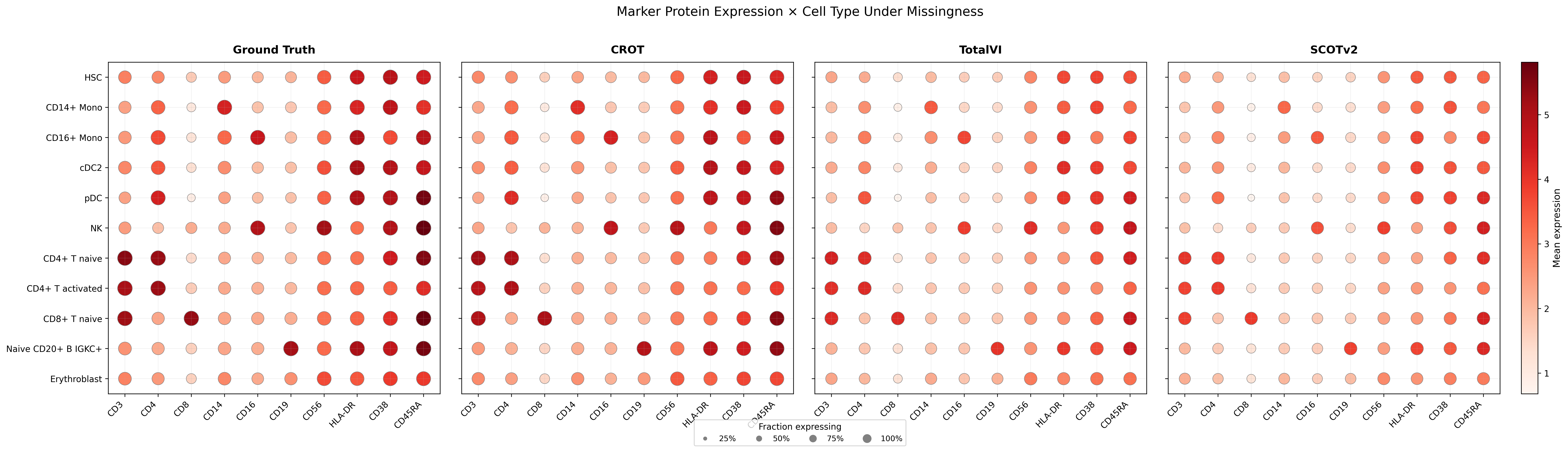}
  \caption{\hl{Dot plots of canonical marker protein expression across cell types in the CITE-seq dataset. Each dot encodes mean expression (color) and fraction of expressing cells (size) for a given marker--cell type pair. Ground truth (leftmost) is compared against CROT, TotalVI, and SCOTv2. CROT most closely reproduces the ground truth marker expression patterns across all cell types.}}
  \label{fig:marker}
\end{figure*}

\hl{To assess whether CROT preserves biologically meaningful signals beyond numerical and clustering metrics, we examine the expression patterns of canonical surface protein markers across cell types in the CITE-seq dataset. We select ten well-established markers---CD3, CD4, CD8, CD14, CD16, CD19, CD56, HLA-DR, CD38, and CD45RA---that are widely used to define major immune cell populations, and visualize their mean expression and fraction of expressing cells per cell type using dot plots.

Figure\mbox{~\ref{fig:marker}} compares the marker expression profiles of the ground truth data against CROT and two competitive baselines (TotalVI and SCOTv2). CROT closely recapitulates the ground truth expression patterns: T cell subsets (CD4+ T naive, CD4+ T activated, CD8+ T naive) show appropriately high CD3 and CD4/CD8 expression, monocyte populations exhibit strong CD14 and CD16 signals, and B cells display elevated CD19, consistent with the expected biological profiles. In contrast, TotalVI and SCOTv2 show notable deviations, including attenuated marker specificity and reduced contrast between cell types. These results confirm that CROT not only achieves accurate numerical recovery but also faithfully preserves the cell-type-specific
marker signatures that are essential for downstream biological interpretation.}

\subsection{Computational Efficiency}
Because the clustering step involves discrete assignments, our optimization follows an alternating scheme where clustering is recomputed at each iteration while gradients are applied only to the continuous imputed variables. Empirically, we observe stable convergence of this procedure across datasets. We compared the training and imputation times of our method against various benchmark methods across three datasets, with results presented in Figure \ref{fig:runnint_time}. Runtime was measured as the total wall-clock time including model initialization, training (if applicable), and inference/imputation on the target dataset. All methods were executed on the same hardware environment. For GPU-enabled methods, GPU acceleration was enabled; CPU-only methods were executed on CPU. As shown in the figure, our method demonstrates a significantly lower runtime compared to other methods, requiring only 5, 12, and 39 seconds for the Cite-seq, Multiome, and PBMC datasets, respectively. On average, our method's runtime constitutes only 1.12\%, 1.29\%, and 2.21\% of the average runtime of the comparison methods on these three datasets. Some baselines require heavy model training (e.g., deep neural networks), while others are iterative interpolation methods. Our reported runtime therefore reflects the complete pipeline cost required to obtain the final imputed dataset.

\subsection{Ablation Study}

\subsubsection{Effectiveness of Clustering Module}

We performed an ablation study to assess the individual contributions of the two loss components in our proposed method, with results presented in Table \ref{tab:ablation}. The findings reveal that optimizing solely for similarity among observed entries already provides strong numerical recovery, but the combination of both losses yields the best overall results, underscoring the effectiveness of incorporating clustering regularization into the imputation process.

\begin{table}[htbp]
\centering
\caption{Ablation study on optimization strategy on CITE-seq. $L_1$ and $L_2$ denote Sinkhorn divergence between data and centroids, respectively. Each value is reported as mean ± standard deviation over five random batch pairs.}
\label{tab:ablation}
\begin{tabular}{lcccc}
\toprule
\textbf{Dataset} & \multicolumn{4}{c}{\textbf{CITE-seq}} \\
\cmidrule(lr){2-5}
\textbf{Metrics} & \textbf{PCC} & \textbf{RMSE} & \textbf{ARI} & \textbf{NMI} \\
\midrule
CROT (w/o $L_2$) & 0.74\pms{0.02} & 1.28\pms{0.05} & 0.78\pms{0.03} & 0.60\pms{0.03} \\
CROT (w/o $L_1$) & 0.71\pms{0.02} & 1.18\pms{0.04} & 0.77\pms{0.03} & 0.70\pms{0.03} \\
CROT (full)      & \textbf{0.76}\pms{0.04} & \textbf{0.99}\pms{0.03} & \textbf{0.82}\pms{0.01} & \textbf{0.75}\pms{0.02} \\
\bottomrule
\end{tabular}
\end{table}

\subsubsection{Improvements over Initialization}

We conduct an ablation study to assess the impact of cluster regularization on our imputation method, as shown in Figure \ref{fig:metrics_evaluation}. In one CITE-seq setting, RMSE, PCC, and ARI metrics initially decline but subsequently improve, ultimately surpassing initial values except for RMSE. This early decline results from the optimal transport algorithm's initial imputed values, which are distant from the true distribution. As optimization advances, the algorithm refines the transport plan, aligning the imputed data with the underlying structure. The final RMSE remains higher due to mean column value initialization, which, though numerically similar, lacks biological relevance compared to the more biologically meaningful patterns captured later.

\subsubsection{Sensitivity Analysis}
\begin{figure}[t]
  \centering
  \includegraphics[width=\columnwidth]{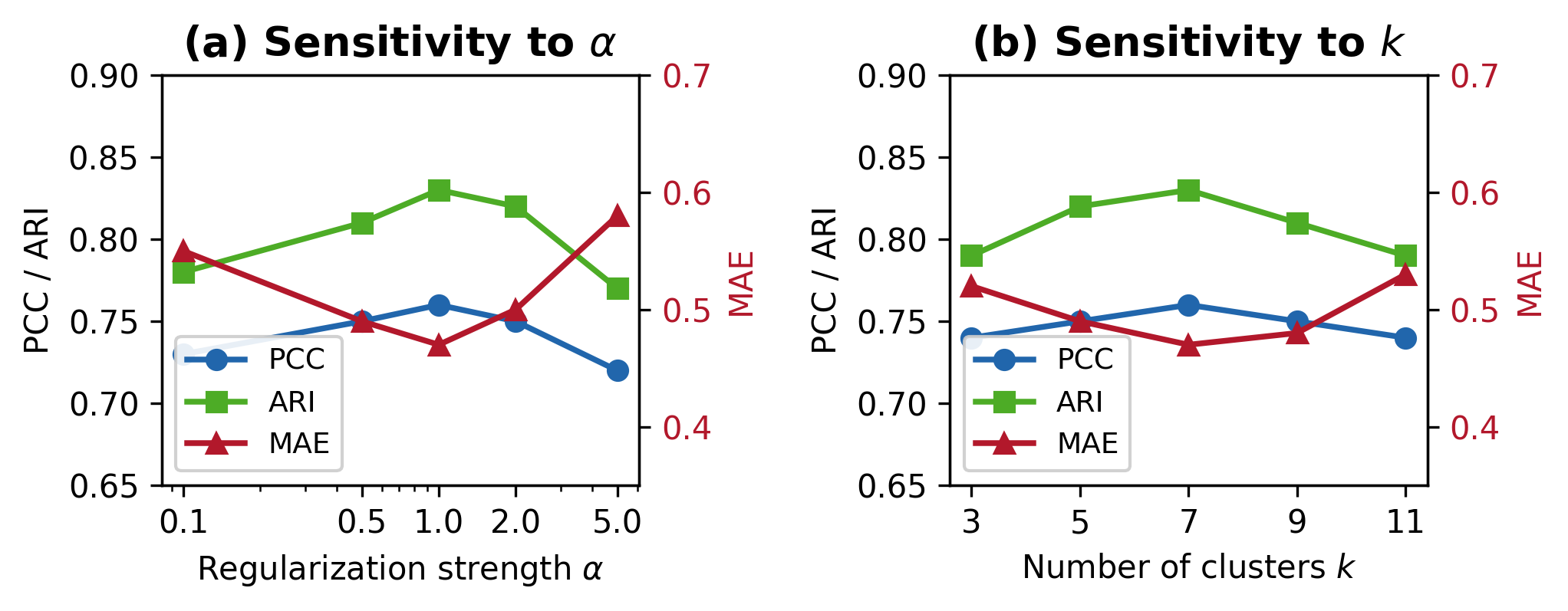}
  \caption{\hl{Sensitivity analysis on CITE-seq. (a)~Performance across varying regularization strength $\alpha$ with $k$ fixed. (b)~Performance across varying number of clusters $k$ with $\alpha = 1$. CROT remains stable across a wide range of both hyperparameters.}}
  \label{fig:sensitivity}
\end{figure}

\hl{We examine the sensitivity of CROT to two key hyperparameters: the cluster regularization strength $\alpha$ and the number of clusters $k$, evaluated on the CITE-seq dataset. As shown in Figure\mbox{~\ref{fig:sensitivity}}(a), performance remains stable across a wide range of $\alpha$ values, with the best results at $\alpha = 1$. Moderate deviations (0.5--2.0) yield comparable performance, indicating that the method is not brittle with respect to the regularization trade-off. Figure\mbox{~\ref{fig:sensitivity}}(b) shows that CROT is similarly robust to the choice of $k$: varying the number of clusters by $\pm 2$ around the Elbow-selected value produces only minor fluctuations in PCC, MAE, and ARI. This confirms that the Elbow-based selection provides a reliable default and that the transferred cluster count from complete to incomplete data does not introduce fragility.}

\subsection{Robustness Analysis}

\hl{To evaluate the robustness of CROT beyond the patch-based setting, we conduct two additional experiments on CITE-seq.}

\subsubsection{Partial Random Missingness}
\hl{Instead of masking an entire modality, we randomly mask 50\% and 75\% 
of the ADT features in the target batch. Table\mbox{~\ref{tab:partial}} shows that CROT achieves the best imputation accuracy at 50\% masking across all three metrics. At 75\% masking, CROT attains the lowest MAE (0.44), while remaining competitive on PCC and RMSE. These results demonstrate that CROT is not restricted to full-modality dropout and degrades gracefully as the missingness ratio increases.}

\begin{table}[t]
\caption{\hl{Results under partial random missingness on CITE-seq. 50\% and 75\% indicate the fraction of ADT features masked in the target batch. 100\% corresponds to the original setting in Table\mbox{~\ref{tab:similarity_sc}}. Each value is reported as mean ± standard deviation over five random batch pairs.}}
\label{tab:partial}
\centering
\small
\resizebox{\columnwidth}{!}{%
\begin{tabular}{l|ccc|ccc}
\toprule
 & \multicolumn{3}{c|}{50\% masked} & \multicolumn{3}{c}{75\% masked} \\
Method & PCC & MAE & RMSE & PCC & MAE & RMSE \\
\midrule
MAGIC   & 0.76\pms{0.02} & 0.94\pms{0.07} & 1.05\pms{0.05} & 0.73\pms{0.04} & 1.01\pms{0.06} & 1.11\pms{0.08} \\
totalVI & 0.78\pms{0.01} & 0.70\pms{0.04} & 0.78\pms{0.08} & 0.75\pms{0.02} & 0.71\pms{0.06} & \textbf{0.77}\pms{0.07}\\
SCOTv2  & 0.80\pms{0.01} & 0.62\pms{0.05} & 1.28\pms{0.18} & \textbf{0.79}\pms{0.01} & 0.75\pms{0.22} & 1.56\pms{0.19} \\
CROT    & \textbf{0.82}\pms{0.01} & \textbf{0.37}\pms{0.01} & \textbf{0.66}\pms{0.02} & 0.77\pms{0.01} & \textbf{0.44}\pms{0.04} & 0.87\pms{0.01} \\
\bottomrule
\end{tabular}%
}
\end{table}

\subsubsection{Mismatched Cluster Composition}
\hl{To evaluate robustness when the cluster structure between reference and target batches differs, we remove all CD14+ monocytes (~37\% of cells) from the target batch in CITE-seq before imputation. This simulates a realistic scenario where the two batches do not share identical cell-type compositions, directly challenging CROT's cluster alignment assumption. Results are shown in Table\mbox{~\ref{tab:mismatch}}. As expected, OT-based methods are more sensitive to this mismatch: SCOTv2 shows notable RMSE degradation (1.99 vs. 1.84 in the standard setting), while CROT also declines from its standard performance. Nevertheless, CROT achieves the lowest MAE (0.79) and remains competitive on other metrics, placing second to totalVI. The stronger robustness of totalVI is consistent with its VAE architecture, which learns a shared latent space without explicit cluster alignment and thus does not assume matched cell-type composition. These results suggest that while CROT's cluster regularization is most effective when cluster structures are similar, it degrades gracefully under moderate mismatch rather than failing catastrophically.}

\begin{table}[t]
\caption{\hl{Results under mismatched cluster composition on CITE-seq. One cell type (CD14+, $\sim$37\% of cells) is removed from the target batch before imputation. Each value is reported as mean ± standard deviation over five random batch pairs.}}
\label{tab:mismatch}
\centering
\small
\begin{tabular}{l|cccc}
\toprule
Method & PCC & MAE & RMSE & ARI \\
\midrule
MAGIC   & 0.70\pms{0.07} & 1.14\pms{0.16} & 1.27\pms{0.21} & 0.57\pms{0.14} \\
totalVI & \textbf{0.73}\pms{0.01} & 0.82\pms{0.05} & \textbf{0.97}\pms{0.14} & \textbf{0.70}\pms{0.09} \\
SCOTv2  & 0.69\pms{0.02} & 0.91\pms{0.04} & 1.99\pms{0.06} & 0.64\pms{0.05} \\
CROT    & 0.71\pms{0.04} & \textbf{0.79}\pms{0.01} & 1.21\pms{0.04} & 0.69\pms{0.02} \\
\bottomrule
\end{tabular}
\end{table}


\section{Conclusion}
We introduced Cluster-Regularized Optimal Transport (CROT), a fast and robust imputation framework designed to address patch-based missingness in single-cell sequencing data. By coupling optimal transport with cluster-level regularization, CROT effectively preserves biological structure while ensuring accurate value recovery. Extensive experiments across multiple datasets demonstrate that CROT achieves superior imputation accuracy and clustering consistency with significantly reduced runtime. These results highlight its practicality for large-scale, multimodal single-cell analysis and its potential as a general solution for structured missing data.

\section{Limitations}

\hl{While CROT demonstrates strong performance across our experimental settings,
several limitations should be noted.}

\paragraph{Batch Effects.}
\hl{First, CROT does not perform explicit batch correction. The optimal transport objective implicitly accommodates moderate batch variation by aligning distributions and cluster centroids between the reference and target data. However, in all our experiments the source and target batches are drawn from the same study and platform, where batch effects are relatively mild. Under severe batch effects, such as cross-laboratory or cross-platform settings, the transport plan may conflate technical variance with biological signal, potentially degrading imputation quality. Incorporating established batch correction methods (e.g.,  Harmony\mbox{~\cite{korsunsky_fast_2019}}, ComBat\mbox{~\cite{johnson_adjusting_2007}})
as a preprocessing step, or integrating batch-aware regularization directly into the transport objective, are promising directions for future work.}

\paragraph{Reference Quality Dependence.}
\hl{CROT assumes that the reference data $X_1$ and the target data $X_2$
share similar biological composition. When this assumption is violated. For
example, when a cell type is absent from the target batch, imputation
quality degrades, as shown in our mismatched cluster composition experiment
(Table~5). More generally, the cluster count $k$ is determined from $X_1$
via the Elbow method and transferred to $X_2$, which may be suboptimal
when the two batches differ in cell-type diversity. Developing adaptive
strategies that allow $k$ to differ between reference and target, or that
jointly infer cluster structure across both, would improve robustness in
heterogeneous settings.}

\bibliographystyle{ACM-Reference-Format}
\bibliography{references}

\newpage
\appendix
\section{Appendices}
\subsection{Data Preprocessing}
\label{sec:preprocessing}
\paragraph{Preprocessing.}
For all three datasets, we follow a standard single-cell preprocessing pipeline using Scanpy~\cite{wolf_scanpy_2018}. Specifically, for each modality we apply library size normalization (\texttt{sc.pp.normalize\_total}) followed by log-transformation (\texttt{sc.pp.log1p}). After normalization, highly variable features are selected: 2{,}000 genes for gene expression (GEX), 4{,}000 peaks
for chromatin accessibility (ATAC), and all available features for protein expression (ADT). CROT does not include a built-in normalization step and operates directly on this pre-normalized input. All baseline methods receive identically preprocessed data.

\paragraph{Dataset Statistics.}
Table~\ref{tab:dataset_stats} summarizes the batch structure and final
feature dimensions of each dataset after preprocessing.

\begin{table}[h]
\centering
\caption{Dataset statistics after preprocessing.}
\label{tab:dataset_stats}
\setlength{\tabcolsep}{3pt}
\begin{tabular}{llrrc}
\toprule
Dataset & Mod. & Cells & Feat. & Sparsity \\
\midrule
\multirow{2}{*}{CITE-seq} & GEX & 90{,}261 & 2{,}000 & 91.9\% \\
                           & ADT & 90{,}261 & 134    & 16.2\% \\
\midrule
\multirow{2}{*}{Multiome}  & ATAC & 69{,}249 & 4{,}000 & 97.4\% \\
                           & GEX  & 69{,}249 & 2{,}832 & 94.5\% \\
\midrule
\multirow{2}{*}{PBMC}      & RNA & 80{,}000 & 1{,}714 & 87.4\% \\
                           & ADT & 80{,}000 & 228    & 8.5\%  \\
\bottomrule
\end{tabular}
\end{table}

\subsection{Baselines}
AutoImpute utilizes autoencoder-based neural networks to handle sparse gene expression matrices by learning the data’s inherent distribution and imputing missing values in scRNA-seq data with minimal modifications to silent gene expressions. MAGIC, another method, employs data diffusion techniques to share information across similar cells, thereby mitigating dropout noise and revealing underlying gene-gene relationships. DeepImpute takes advantage of deep neural networks, leveraging dropout layers and loss functions to predict missing values in scRNA-seq data, with a focus on scalability and efficiency. For multimodal imputation, JAMIE uses joint variational autoencoders, integrating data from different modalities by learning shared latent spaces, which can then be used for cross-modal predictions. ScButterfly extends this approach by incorporating dual-aligned variational autoencoders with data augmentation to facilitate cross-modality translation, specifically targeting scenarios with unpaired or noisy multimodal data. scBFP introduces a two-step graph-based feature propagation for scRNA-seq imputation, combining gene-gene and cell-cell relationships to improve data denoising and imputation accuracy. totalVI jointly models RNA expression and surface protein abundance using a hierarchical probabilistic framework based on variational inference. It explicitly accounts for technical noise in protein measurements and learns a shared latent representation across modalities, enabling cross-modality prediction and data integration. MultiVI extends the variational inference framework to integrate gene expression and chromatin accessibility measurements. By learning a shared latent representation of cells across modalities, MultiVI enables modality translation and missing-modality imputation. SCOTv2 is an optimal transport–based alignment framework that employs Gromov-Wasserstein distance to align multimodal single-cell datasets while preserving intrinsic geometric structures within each modality. It provides a scalable solution for cross-modality mapping and imputation.

\subsection{Alignment of Observed and Imputed Data}
\begin{figure*}[htbp]
    \centering
    \includegraphics[width=0.98\textwidth]{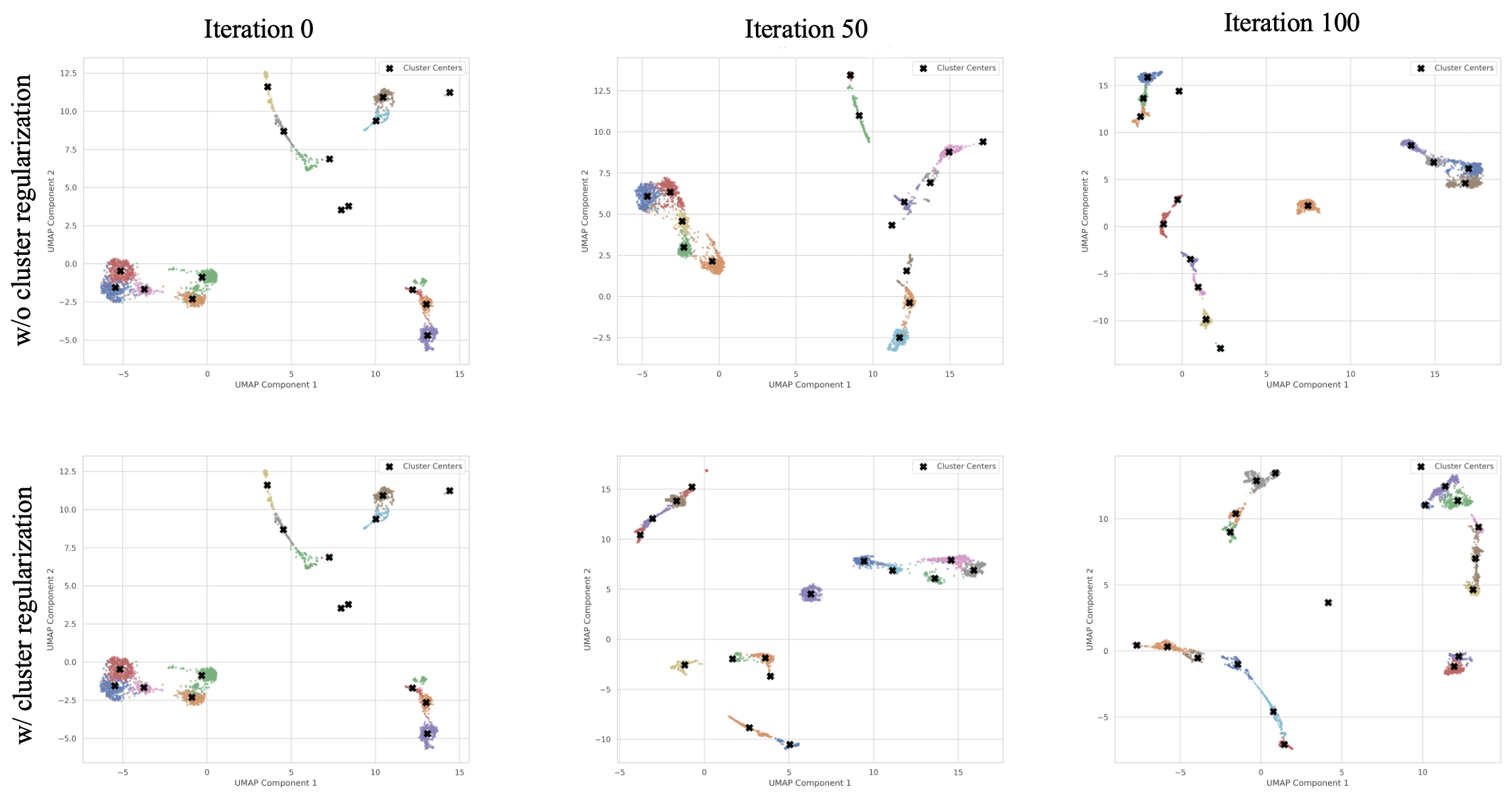}
    \caption{UMAP visualizations showing clusters of different iterations during imputation. Better distinct clusters of UMAP plots helps with further analysis of single-cell data, such as the presence of rare cell types or intermediate states between well-defined populations.}
    \label{fig:umap2}
\end{figure*}

We plotted umap graphs at different stages of interpolation. As shown in the Fig \ref{fig:umap2}, the class separation effect after using cluster loss is slightly better than that without using it. Our effect is not obvious yet, and we expect to further refine the class constraint strategy in the future.

\subsection{Evaluation of Global Structure Consistency}
\begin{figure*}[htbp]
\centering
\begin{minipage}{0.5\textwidth}
    \centering
    \includegraphics[width=\linewidth]{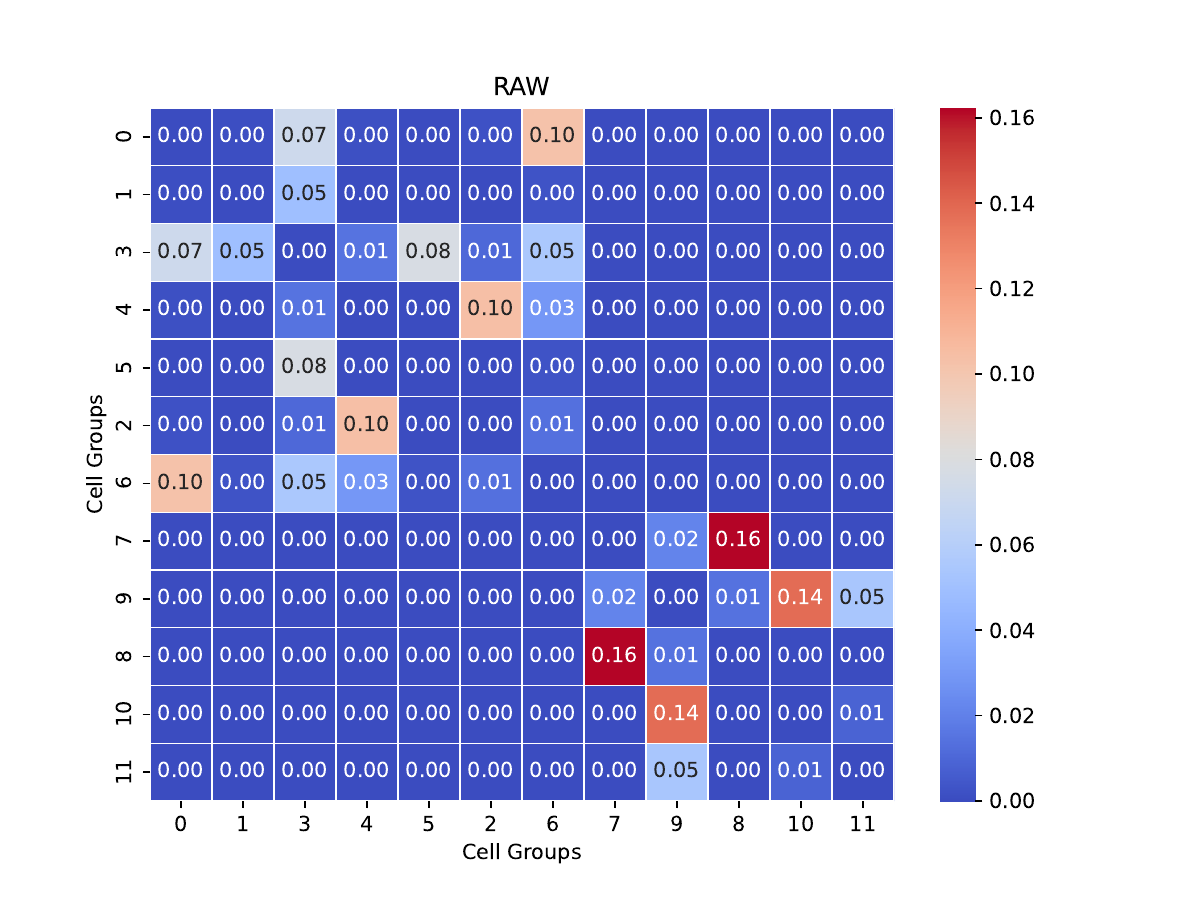} 
\end{minipage}%
\begin{minipage}{0.5\textwidth}
    \centering
    \includegraphics[width=\linewidth]{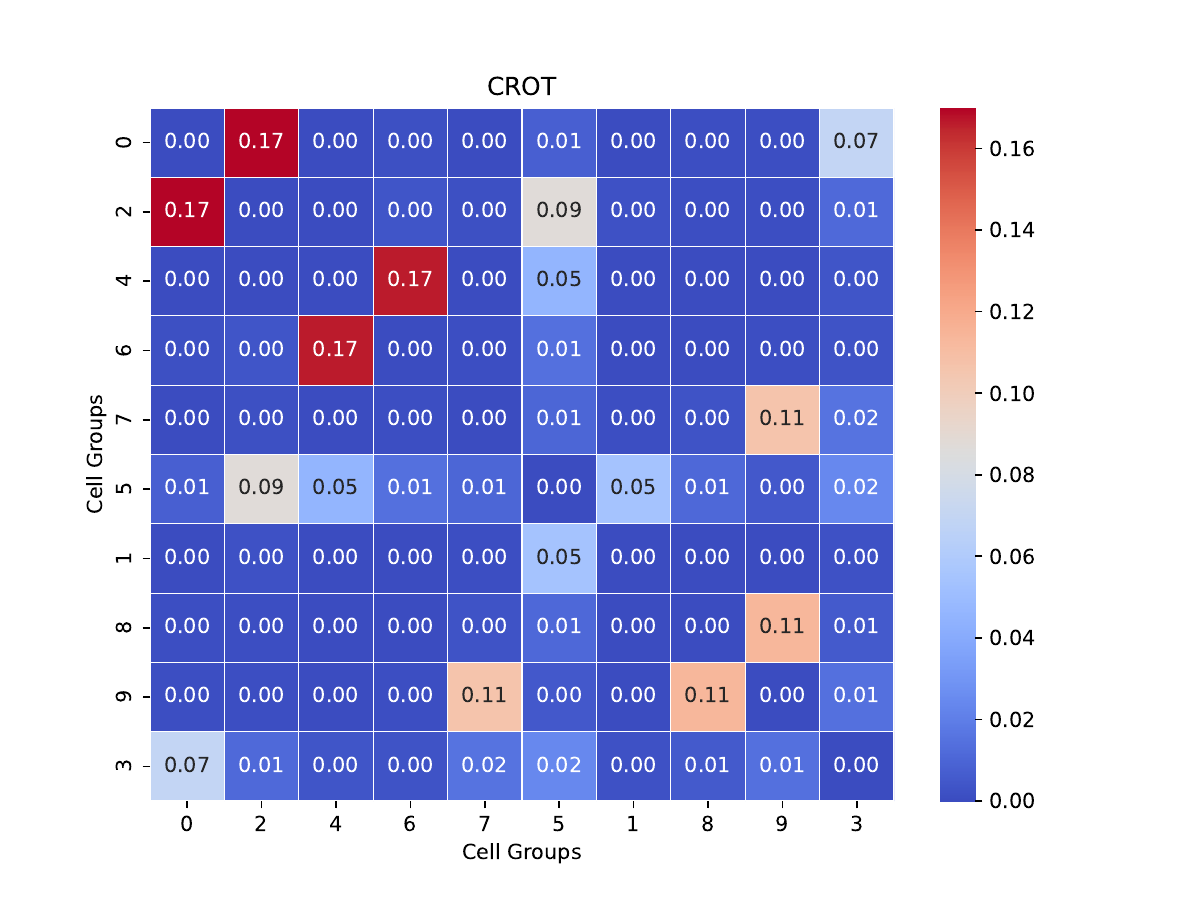}
\end{minipage}

\caption{PAGA similarity heatmap of RAW and imputed data.}
\label{fig:paga}
\end{figure*}

This section presents the PAGA similarity heatmap (Fig \ref{fig:paga}), a quantitative representation of the global connectivity structure among cell clusters derived from the single-cell RNA-seq data. The heatmap highlights the pairwise connectivity strengths between clusters, computed using the PAGA algorithm. Each value represents the normalized connectivity, reflecting the similarity between clusters based on their transcriptional profiles.

We use the PAGA similarity heatmap to evaluate the global structural consistency before and after data imputation. By comparing heatmaps generated from raw (missing) and imputed datasets, we demonstrate how our proposed imputation method restores biologically meaningful cluster relationships. 

\subsection{Implementation Details}
Since our method is relatively simple, it does not require a lot of parameter tuning. We think that increasing the batch size for each sampling may help, but given the resource constraints, our current choice ($\text{batch}\_\text{size}=3000$) has performed well enough.

We follow the suggested hyperparameter settings by the authors:

\subsubsection*{DeepImpute\footnote{\url{https://github.com/lanagarmire/deepimpute}}}
We loaded two modalities of observed data as \texttt{data1}, \texttt{data2}, then initialized 5 JAMIE models with the epoch number in \{300, 500, 700, 1000, 1500\}, since the authors suggested 500 as their setting, and our data was larger. The other parameters were set as default. After training the initial model, we chose the best one and performed data imputation using observed entries.

\subsubsection*{AutoImpute\footnote{\url{https://github.com/divyanshu-talwar/AutoImpute}}}
Since AutoImpute utilizes \texttt{.csv} files, we first translated our data into \texttt{.csv} format. The key hyperparameter is \texttt{hidden\_units}, which controls the size of the hidden layer or latent space dimensions, and we set \texttt{hidden\_units=3500} instead of the default value of 2000.

\subsubsection*{scBFP\footnote{\url{https://github.com/Junseok0207/scBFP}}}
We found \texttt{gene\_k} and \texttt{cell\_k} = 40, \texttt{gene\_iter} and \texttt{cell\_iter} = 100, provided the best performance.

\subsubsection*{DeepImpute\footnote{\url{https://github.com/lanagarmire/deepimpute}}}
We adjusted \texttt{limit} (Genes to impute, e.g., first 2000 genes. Default: \texttt{auto}) to equal the number of missing entries in each dataset, and conducted experiments under \texttt{max-epochs} in \{300, 500, 700, 1000\}. Our results showed that \texttt{max-epochs=700} performed best.

\end{document}